\pdfoutput=1

\documentclass[11pt]{article}

\usepackage[]{acl}

\usepackage{times}
\usepackage{latexsym}
\usepackage{subcaption}
\usepackage[T1]{fontenc}

\usepackage[utf8]{inputenc}

\usepackage{microtype}
\usepackage{inconsolata}
\usepackage{booktabs}
\usepackage{multirow}
\usepackage{graphicx}
\usepackage{longtable}
\usepackage{ragged2e}
\usepackage{comment}
\usepackage{tcolorbox}
\usepackage{amsmath}
\usepackage{amssymb}
\usepackage{amsfonts}
\usepackage{bbm}
\usepackage{float}
\title{Epistemology of Language Models: \\Do Language Models Have Holistic Knowledge?}

\author{Minsu Kim, James Thorne\\
Korea Advanced Institute of Science and Technology (KAIST)\\
Seoul, South Korea\\
\texttt{\{minsu\_kim, thorne\}@kaist.ac.kr}}

\begin{document}
\maketitle
\begin{abstract}
This paper investigates the inherent knowledge in language models from the perspective of epistemological holism. The purpose of this paper is to explore whether LLMs exhibit characteristics consistent with epistemological holism. These characteristics suggest that core knowledge, such as general scientific knowledge, each plays a specific role, serving as the foundation of our knowledge system and being difficult to revise. To assess these traits related to holism, we created a scientific reasoning dataset and examined the epistemology of language models through three tasks: Abduction, Revision, and Argument Generation. In the abduction task, the language models explained situations while avoiding revising the core knowledge. However, in other tasks, the language models were revealed not to distinguish between core and peripheral knowledge, showing an incomplete alignment with holistic knowledge principles. 
\end{abstract}

\section{Introduction}

Recent advancements in language models have extended their capabilities beyond simple question-answering (QA) tasks to more complex knowledge-intensive retrieval and reasoning challenges akin to problems solved by humans \citep{petroni2021kilt,lewis2021retrievalaugmented, wei2023chainofthought}. Moving beyond these QA tasks, large-scale language models are now recognized for their ability to revise and update their knowledge, including the implicit consequences of such modifications \citep{zhong2023mquake,cohen2023evaluating}. Furthermore, there is an increasing interest in applying language models in more applied fields such as medicine and law, given their human-like functionality in various knowledge evaluation tasks \citep{inproceedings,henderson2022pile, singhal2023expertlevel,bommasani2023foundation}. 

\begin{figure}[hbt!]

  \centering

  \includegraphics[width=\columnwidth]{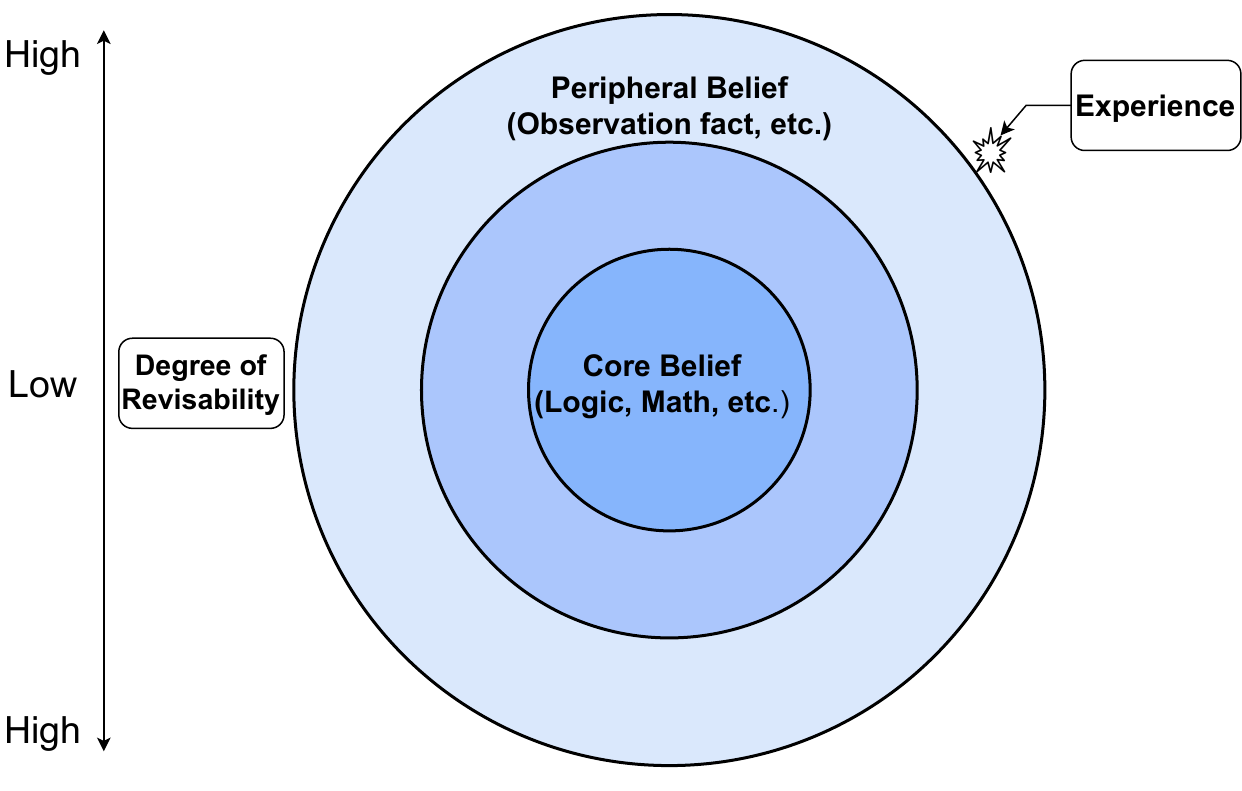}

 \captionsetup{}

  \caption{A diagram of the holistic web of belief. At the core, there are certain pieces of knowledge that serve as the basis of our beliefs. while towards the periphery, less certain empirical knowledge is located. In this web, all knowledge is revisable, but when we encounter new experiences, the peripheral knowledge is more prone to revision than that at the core.}

  \label{fig:brief}

\end{figure}

This paper focuses on a more philosophical inquiry: we explore the nature of the epistemology inherent in language models, questioning what they know, how they process their beliefs, and how they handle knowledge from philosophical viewpoints. 
The primary aim of this paper is to examine the knowledge within language models through the lens of epistemological holism, particularly focusing on whether these models possess \textit{core knowledge}, such general scientific knowledge and commonsense, as outlined by epistemological holism.

One of the tenets of epistemological holism is that our knowledge forms a ``Web of Belief.'', which means that no piece of knowledge is isolated; rather, each is interconnected with others, forming a network. At the core of this network lie relatively certain knowledge elements such as logic, commonsense, and scientific facts, which are difficult to revise even when counterexamples are presented. On the web's periphery are empirical facts, more directly related to experience and more easily revised upon encountering counterexamples \citep{Quine1951-QUITDO-3,Quine1970-QUITWO-2}. Epistemological holism adopts a kind of pragmatic point of view regarding knowledge revision. When faced with a counterexample to a general fact, one tends to defend the general fact as much as possible while seeking alternative conditions or explanations. A classic example is the response to the failure of Newton's laws to precisely predict Uranus's orbit. Instead of revising Newtonian mechanics, scientists hypothesized the existence of another influencing planet, leading to Neptune's discovery \citep{Kuhn1962-KUHTSO-3}.

We developed a dataset based on the World Tree corpus \citep{xie-etal-2020-worldtree} by following the procedure used in the Worker-AI collaboration framework \citep{liu2022wanli}, 
Through this dataset, we evaluated how language models respond to collisions with general knowledge by evaluating three tasks: Abduction, Revision, and Argument Generation, along with fine-tuning.
In the abduction task, all models achieved over 60\% without tampering with core knowledge. However, in revision and argument generation tasks, even state-of-the-art models frequently negated core knowledge. Results from Supervised Fine-tuning also revealed that language models tend to treat general factual knowledge and core knowledge equally. These mixed outcomes imply that LLMs' alignment with a holistic epistemological framework may be context-dependent, indicating a partial, rather than comprehensive, adherence to holistic epistemology. 

\section{Related Works}

\paragraph{Knowledge Editing}
Previous research on knowledge editing has primarily focused on the parametric knowledge update process. \citet{decao2021editing,jang2023temporalwiki} argued the importance of factual knowledge updates in language models as information that changes over time. \citet{mitchell2022fast,meng2023locating,meng2023massediting} introduced methods for updating specific networks to efficiently inject new knowledge. \citet{zheng2023edit} showed that knowledge can be revised through in-context learning (ICL). 
Since the update of a single sentence can impact related sentences, \citet{cohen2023evaluating,zhong2023mquake} have studied the implicit knowledge editing that occurs when knowledge edits happen. \citet{qian2023merge} 
demonstrate that language models are sensitive to external knowledge that conflicts with parametric knowledge. Our paper includes a task related to knowledge editing. However, our aim goes beyond examining the consistency of language models given external facts. We seek to understand how LLMs modify existing knowledge or defend it when presented with abnormal counterfactuals.

\paragraph{Inference Ability of Language Models}
Historically, significant efforts have been dedicated to evaluating and enhancing the reasoning capabilities of language systems. \citep{mccarthy59a}. For example, the Natural Language Inference (NLI) task involves classifying the entailment relationship between two sentences \citep{bowman2015large,williams-etal-2018-broad,nie-etal-2020-adversarial,liu2022wanli}. 
\citet{DBLP:conf/iclr/BhagavatulaBMSH20} introduced a reasoning dataset that evaluates if a language model can provide the best explanation given a scenario.  \citet{zhao2023uncommonsense} introduced research that explains situations where unexpected events occur in everyday life. This paper explores the preference and reasoning of language models in situations where they conflict with general facts and could potentially disrupt our belief system and where multiple indeterministic yet valid logical conclusions exist.

\paragraph{Science Knowledge of Language Models}
The exploration of whether LLMs possess scientific knowledge and are capable of scientific reasoning has also been a subject of research. SCINLI is an NLI dataset with a focus on scientific topics \citep{sadat-caragea-2022-scinli}. Datasets like SCITAIL \citep{Khot2018SciTaiLAT}, ARC \citep{clark2018think}, WorldTree \citep{williams-etal-2018-broad}, OpenBookQA \citep{mihaylov2018suit}  involve school level science problems and evaluate whether language models can find correct answers. As WorldTree dataset includes a dataset involving one sentence scientific facts, we utilized it to make our  $(\text{scientific fact},\text{counter-example},\text{explanations})$ tuple dataset. 

\section{Primer on Epistemology}
Epistemology is a branch of philosophy that explores knowledge. It investigates the conditions under which we can claim to ``know'' something. \textit{Traditional} epistemology has focused on the normative conditions of knowledge, defining knowledge as ``justified true belief'' and contemplating what constitutes justification, the definition of truth, and the nature of belief. While traditional epistemology deals with the normative aspects of knowledge, i.e., the necessary and sufficient conditions of knowledge, \textit{naturalized} epistemology discusses how our knowledge is actually formed and undergoes revision, and holism is one of the most well-known branches of naturalized epistemology \citep{Quine1968-QUIEN,Audi1997-AUDEAC-3}. In this paper, we study the epistemology of language models.

\subsection{Epistemological Holism}
Contrary to the traditional approach that examines propositions in isolation to determine their justification and truth, \citet{Quine1951-QUITDO-3} suggests that no piece of knowledge is isolated but is interconnected with other knowledge in the web of beliefs. When a proposition is tested, it brings the entire related knowledge and knowledge system to the test bench. For instance, in the case of scientific knowledge or theoretical propositions, a single proposition is not tested, confirmed, or refuted alone. It is tested along with other related theories, common sense, and empirical conditions. When our system of knowledge operates in this way, if we encounter observational fact that support the common, accepted scientific facts, it strengthens our belief in those facts. However, if an observation contradicts general knowledge, we must revise our web of belief. 
However, from a pragmatism viewpoint, core propositions such as logical, mathematical, or commonsense knowledge are those we are reluctant to modify as core statements form the basis of our knowledge and are coherently connected with many other beliefs. Rather, we prefer to revise other conditions or empirical observations related to them, and knowledge evolves in this manner \citep{Quine1968-QUIEN, Quine1970-QUITWO-2, Quine1951-QUITDO-3}. The argument of holism in the philosophy of science is illustrated in Figure \ref{fig:Duhem-Quine}.

\begin{figure*}[hbt!]

  \centering

  \includegraphics[width=\textwidth]{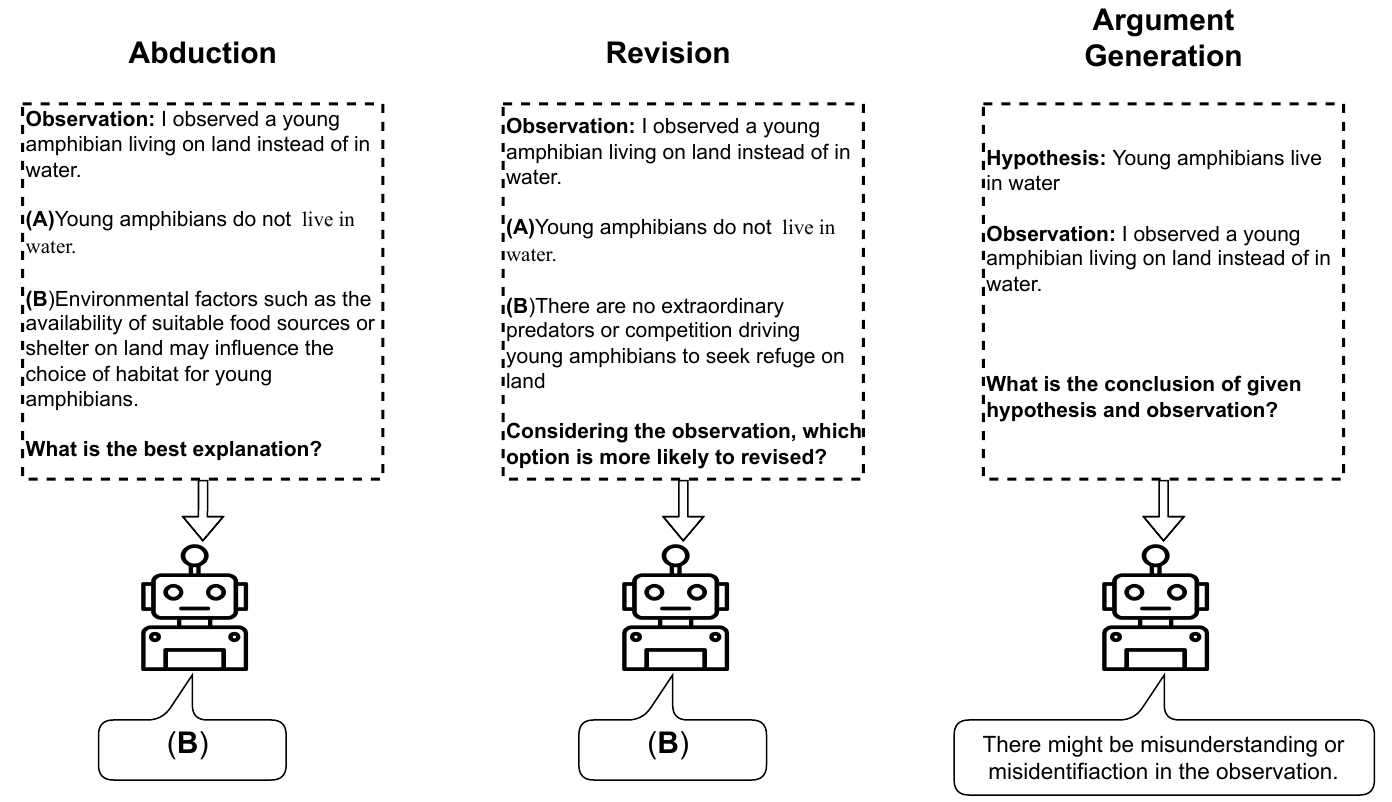}

  \captionsetup{}

  \caption{Introduction of three main tasks. The abduction task is a preference task that seeks to investigate whether LLMs favor abductive explanations over negating core statements. The Argument generation task aims to explore the capability of language models to produce holistic arguments. The revision task is designed to find out whether language models, when faced with counterexamples, prefer to modify peripheral knowledge or instead opt to alter core knowledge. }
  \label{fig:main_task}

\end{figure*}

\section{Dataset}

We construct a holism reasoning dataset based on scientific facts in the WorldTree V2 dataset \citep{xie-etal-2020-worldtree}. Therefore, all scientific facts in our dataset are included in WorldTree V2. We augment the dataset with counterfactual observations. The generation process of the counterfactuals is similar to the one described in \citet{liu2022wanli}. We followed three steps to create our dataset: 
\begin{enumerate}

    \item We initially collect a subset of scientific facts from the WorldTree V2 dataset.
    \item For each fact, we overgenerate several counter-observations and plausible explanations using GPT-3.5-Turbo. 
    \item We select the best explanations and create a tuple of the scientific fact, counter observation, and possible explanations. 
\end{enumerate}
We represent our sample as quadruplet $(s,c,e^{*}_{1},e^{*}_{2})$ where $s$ is general scientific fact, $c$ is counter observation, $e^{*}_{1},e^{*}_{2}$  are plausible explanations.

\subsection{Extraction}
The WorldTree dataset also released a table $W$ for creating explanations for scientific problems, which stores one sentence of scientific knowledge. We derived scientific facts from the given store $W$ and made a new scientific knowledge store $W^{'}  \subset  W$. Not all facts were extracted; those with difficult-to-create counterexamples and possible conditions were filtered out. For example, we excluded scientific definitions, tautological sentences, sentences more logical than scientific, vague or ambiguous statements, and sentences closer to ethics than science. 

\subsection{Overgeneration}
While creating the dataset, we leveraged the few-shot ability of a language model $F$ by providing demonstrations that include negation of scientific facts and mention of other conditions, compelling the models to generate explanations.  $F(s_{i},d)=(s_{i},c_{i},e_{i1},e_{i2},e_{i3},e_{i4},e_{i5})$ where $s_{i} \in W^{'}$, $d$ is demonstration examples, $c_{i}$ is generated counter example, and $e_{i1} \sim e_{i5}$ are generated possible explanations. The generated counter examples $c_{i}$ start with an observation situation prefixes such as ``I observed the fact'' or ``I discovered that''. We generate observation statements that contradict the general scientific fact $s_{i}$. For example, if the given scientific fact is ``sharks live in oceans.'', the counter observation might be ``I observed a shark living in a freshwater lake.''  For explanations, similar to \citet{liu2022wanli},  we overgenerated,  producing several explanations by using the language model. The first explanation $e_{i1}$ is a direct negation of the scientific fact $s_{i}$. For example, if the fact is ``sharks live in oceans,'' the model generates ``sharks do not live in oceans.'' In other words, it essentially creates a hasty generalized statement. The second explanation $e_{i2}$ uses negation with 'some,' like ``some sharks do not live in the ocean.'' The remaining third to fifth candidate explanations $e_{i3} \sim e_{i5}$ were freely generated by the model based on in-context learning. 
We augmented data using GPT-3.5 with a 4-shot demonstration.

\subsection{Filtering and Modification}

After the explanations for a fact were generated from a language model, yielding $s_{i},c_{i},e_{i1},e_{i2},e_{i3},e_{i4},e_{i5}$, a filtering process was performed. During the filtering process, we either used the generated responses verbatim from the language model or manually modified them or deleted the sample if the language model just repeat the input. For example, if the counterexample does not start with ``I observed'' but simply negates the general fact, we added the prefix manually. Then, we select a single explanation among those, excluding the first one that plausibly explained the counterexample. The generated counterexamples and explanation candidates tended to have low diversity. Therefore, during the filtering process, we manually selected a variety of explanations for the dataset to ensure that they do not overlap. We did not verify these explanations for their scientific validity and verifiability. These explanations could contradict actual scientific facts and be fictional conditions. However, we allowed this as speculative imagination is often a part of scientific inquiry. Our final dataset can be represented as $H$ containing quadruplet $(s,c,e^{*}_{1},e^{*}_{2})$ where $s$ is general scientific fact, $c$ is counter observation, $e^{*}_{1},e^{*}_{2}$  are filtered explanation where $e^{*}_{1}$ is direct negation of $s$ and $e^{*}_{2}$ is abductive explanation for  the given contradicting situation $(s,c)$. After filtering, we were left with 1,709 samples, the statistics are reported in the appendix in Table \ref{tab:data_statistics}.

\section{Task}

Figure~\ref{fig:Duhem-Quine} shows the holistic argument in the philosophy of science, called Duhem-Quine thesis \citep{Duhem1954-DUHTAA}. 
The motivation and hypothesis behind our proposed task are as follows: Holism assigns a special role to core statements, distinct from other empirical propositions. Therefore, if language models possess a holistic knowledge network when they encounter facts that contradict the core statements, they will avoid altering the core statements as much as possible.

Based on this argument and motivation, we have developed three tasks to assess holism in language models: abduction, revision, and argument generation. You can see the examples of each task in Figure \ref{fig:main_task}

\paragraph{Abduction}
The first task is to determine the preference of a language model for two explanations. Given the inputs $(c, e^{*}_{1},  e^{*}_{2})$, we assess if a language model prefers $e^{*}_{1}$, which negates a general scientific fact, or $e^{*}_{2}$, which mentions a specific condition that protects the general scientific fact and explains $c$.  We do not explicitly give $s$ representing a general fact in the input, as it is assumed that this information is stored in the model's parametric knowledge as a core belief \citep{petroni2019language}. If the language model has a holistic system, it will choose $e^{*}_{2}$ to avoid simply negating the general fact. The formula for the task is as follows:

{
\small
\begin{equation}
\begin{split}
F((c, e^{*}_{1},  e^{*}_{2})) = \begin{cases} 
e^{*}_{1} & \text{if } F \text{ directly negates } s \text{ to explain } c, \\
e^{*}_{2} & \text{if } F \text{ prefers to protect } s \text{ and uses} \\
& \quad \text{a specific condition to explain } c.
\end{cases}
\end{split}
\end{equation}
}

\renewcommand{\arraystretch}{1.5}
\begin{figure}[t!]
\centering
\scriptsize
\[
\begin{array}{cl}
\textbf{Hypothesis:}&\text{Scientific fact} \land \\
& (\text{Auxiliary hypothesis} \land \text{Observation conditions}) \\
\textbf{Observation:}&\text{Observed fact that can refute Scientific fact} \\
\hline
\textbf{Conclusion:}&\neg \text{Scientific fact} \lor \\
& \neg \text{Auxiliary hypothesis} \lor \neg \text{Observation conditions} \\
\end{array}
\]
\caption{The argument of Duhem–Quine thesis. In the hypotheses, implicit assumptions are interconnected with explicit scientific facts. When an observation contradicts a scientific fact, it challenges both the fact and related statements. The conclusion of this process is indeterministic, meaning we can either negate the scientific fact or other implicit propositions. However, as most scientific facts are the basis of our web of belief, we often end up negating auxiliary hypotheses or observational conditions. }
\label{fig:Duhem-Quine}
\end{figure}

\paragraph{Revision}
The revision task is a knowledge editing task. Similar to the abduction task, when a counterexample is provided, this task involves choosing which option should be modified in the given counterexample. Specifically, the language model receives $(c, s, \neg e^{*}_{2})$ as input. This means for the counter observation $c$, the first candidate for modification is the general fact $s$, and the second candidate for modification is the negation of $e^{*}_{2}$, which denies the existence or effect of a specific condition or the presence of a measurement error. For example, this task involves deciding which statement to modify when the language model encounters the sentence, $c=$ ``I observed that there is a celestial object that is closer to the Earth than the moon.'' The choices are between modifying $s=$ ``the moon is the celestial object that is closest to the Earth'' or $\neg e^{*}_{2}=$ ``There are \textit{no} extraordinary near-Earth asteroids or comets passing by our planet in this circumstance.''  According to epistemological holism, ideally, the language model is likely to be reluctant to modify the core statement $s$. Therefore, it will select  $\neg e^{*}_{2}$, thereby protecting its system of knowledge. The formula for the task is as follows:

{
\small
\begin{equation}
\begin{split}
F((c, s, \neg e^{*}_{2})) = \begin{cases} 
s & \text{if } F \text{ directly negates } s \text{ to explain } c, \\
\neg e^{*}_{2} & \text{if } F \text{ protects } s \text{ and prefers to} \\
& \quad \text{revise } \neg e^{*}_{2} \text{ to explain } c.
\end{cases}
\end{split}
\end{equation}
}

\paragraph{Argument Generation}
In this task, a hypothesis $s$ and an observation $c$ are provided to the language model, and the model is asked to infer a conclusion from this context. Similar to the formula in figure \ref{fig:Duhem-Quine}, we conducted controlled hypothetical inference where auxiliary theories/hypotheses or specific conditions were not given. This is because the essence of this task is to uncover these latent conditions or to challenge explicit hypotheses. If the language model exclusively draws conclusions that negate the hypothesis $s$, it suggests that the language model does not consider general facts or commonsense as core facts within a holistic system, contrary to what is expected in epistemological holism. The formula for the task is as follows:

{
\small
\begin{equation}
\begin{split}
F((s,c)) = \begin{cases} 
\neg s & \text{if } F \text{ directly negates } s \text{ to explain } c, \\
e^{'} & \text{if } F \text{ protects } s \text{ and uses} \\
& \quad \text{a specific condition to explain } c \text{ and } s.
\end{cases}
\end{split}
\end{equation}
}
where $e^{'}$ is a generated sentence that explains the given observation c without altering  $s$.

\section{Experimental Design}
We conducted evaluations for the aforementioned tasks using the FLAN-T5 models (ranging from base to XXL) \citep{chung2022scaling}, Llama2-chat-7b and 13b \citep{touvron2023llama}, Phi-2 \citep{gunasekar2023textbooks}, GPT-3.5, and GPT-4 \citep{openai2023gpt4}. All tasks were executed as zero-shot evaluations. The reason for not employing in-context learning is that the purpose of this paper is not to enhance the performance of each task but to investigate whether language models consider scientific facts and other universal truths as part of their core knowledge that is difficult to revise. This is because, with a few-shot demonstration, language models can easily choose between two options based on sentence patterns. We used a total of 1,709 data samples as test samples for zero-shot evaluation. 
\paragraph{Evaluation Metric}
For convenience, we will refer to general scientific knowledge as core knowledge and statements or knowledge that explains the conflicting situation by mentioning other conditions or factors or raising questions about observation as peripheral statements. 
We measured the metric  Peripheral Response Ratio (PRR), showing the proportion of instances out of the total samples where the model explained the conflicting situation using peripheral statements rather than directly negating or modifying general facts. 

\begin{equation}
\text{PRR} = \frac{\text{The Number of peripheral statements}}{\text{Total number of samples}}
\end{equation}
Hence, in abduction or revision tasks, PRR refers to the accuracy of the model choosing the peripheral statement; in the argument generation task, PRR refers to the accuracy of the model response, including the peripheral statement.
For all evaluations, we generated one sentence through greedy decoding. 

\section{Results and Discussion}

\paragraph{Preference Tasks: Do language models prefer to keep general knowledge?}

In the abduction task, all the models used in the experiment showed a preference for explaining given atypical observations by mentioning the peripheral statements rather than negating a general fact. Particularly in the case of the Flan models, it was observed that the PRR increased as the model size grew, with Phi-2 showing 90\%  PRR. The type of GPT models showed approximately 80\% PRR for the abduction task.

However, in the revision task the language models, in contrast to the abduction task, showed a low PRR score. This low PRR implies that when faced with atypical situations, language models prefer to modify core knowledge, such as scientific facts and commonsense. For example, when presented with the observation, ``I observed that the kidney is not located in the abdomen,'' language models favored revising the knowledge ``Kidney is located in the abdomen'' over asserting ``There is no misidentification or mistake in observation.'' (This example is an actual case from GPT-4). Intriguingly, for Flan models, PRR decreased as the model size increased, and only Phi-2, among the decoder models that performed well in abduction tasks, also showed good performance in the revision task. This outcome reveals a different aspect of reasoning in language models.
\begin{table}[t!]

\setlength{\tabcolsep}{15pt}
\resizebox{\columnwidth}{!}{%
\begin{tabular}{@{}lccc@{}}
\toprule
\multirow{2}{*}{\textbf{Model}} & \multicolumn{3}{c}{\textbf{Peripheral Response Ratio (\%)}}                                                                       \\ \cmidrule(l){2-4} 
                                                 & \textbf{Abduction} & \textbf{Revision} & \textbf{\begin{tabular}[c]{@{}c@{}}Argument \\ Generation\end{tabular}} \\ \midrule
Flan-T5-Base                                     & 61.0                              & 81.3                             & 6.0                                         \\
Flan-T5-Large                                    & 66.2                              & 48.9                             & 12.5                                        \\
Flan-T5-XL                                       & 81.0                              & 34.2                             & 10.5                                        \\
Flan-T5-XXL                                      & 83.1                              & 7.1                              & 9.90                                        \\
Llama2-7b-chat                                   & 62.1                              & 47.6                             & 51.8                                        \\
Llama2-13b-chat                                  & 64.3                              & 42.8                             & 38.4                                        \\
Phi-2                                            & 90.9                              & 62.0                             & 17.0                                        \\
GPT-3.5-turbo                                    & 83.7                              & 23.2                             & 15.0                                        \\
GPT-4                                            & 79.5                              & 15.6                             & 32.5                                        \\ \bottomrule
\end{tabular}
}

\caption{PRR (Peripheral Response Ratio) represents the ratio at which the language model negates or modifies knowledge located in the periphery instead of negating or revising the core knowledge. }
\label{tab:main_result}
\end{table}

\begin{figure*}[t!]

  \centering

  \includegraphics[width=\textwidth]{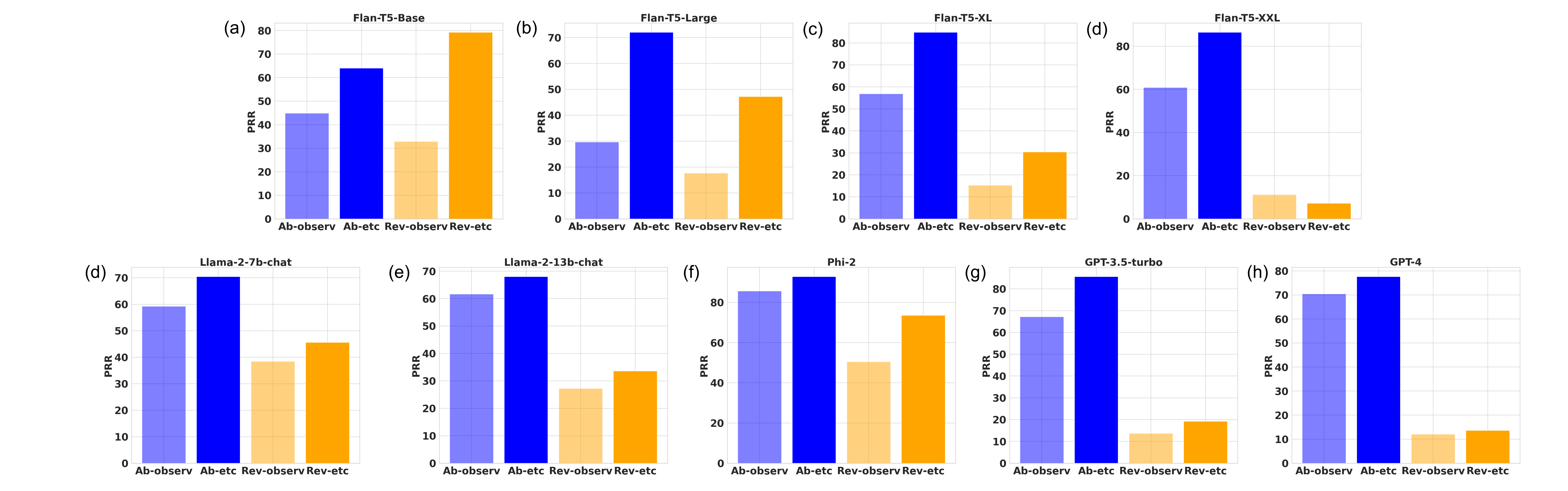}

 \captionsetup{}

  \caption{"Ab\_observ" involves a comparison between negating a general fact and negating an observation fact. On the other hand, "Ab\_etc" contrasts the negation of a general fact with the utilization of other peripheral facts. "Rev\_observ" is a task that involves deciding which needs to be modified between a general fact and the claim that an observation is valid. "Rev\_etc", on the other hand, is a task that determines what needs to be revised between a general fact and the absence of other hypothetical conditions.}

  \label{fig:observ_etc}

\end{figure*}

The primary goal of the abduction task is to select the most plausible explanation for a given observation that fits within the existing knowledge framework. Due to the abundance of instructions for explanations and datasets for abduction, language models are likely to be familiar with these types of tasks \citep{zhao2023uncommonsense}. Also, the task implicitly encourages the model to use peripheral or auxiliary information to construct explanations without directly challenging the core knowledge. The success of LLMs in abduction tasks suggests they are adept at navigating their extensive pre-trained knowledge to find and apply relevant peripheral information that can explain new observations without directly confronting or needing to alter core knowledge. 


Revision tasks differ from abduction tasks by requiring language models (LLMs) to evaluate the validity of core versus peripheral knowledge in the face of contradictions. While abduction tasks allow for generating explanations that work around core knowledge and incorporate additional, compatible information to explain an observation, revision tasks demand a direct assessment of whether to modify core knowledge or adjust peripheral details. The fact that LLMs do not exhibit a holistic approach in revision tasks indicates they may not inherently prioritize protecting core knowledge when faced with its potential revision.

\paragraph{Qualitative Analysis for Argument Generation Task: Can language models make a holistic inference?}
In the argument generation task, which best represents the indeterministic argumentative structure of epistemological holism, we similarly measured the ratio of conclusion where LLMs refer to peripheral statements. 

When measuring the PRR in the generation task, we used a soft criterion. Specifically, as long as the hypothesis was not completely negated, we considered the language model to have made a peripheral response; also, any mention of a peripheral statement was given a positive PRR score. This is because, especially with models like those in the GPT series and Phi-2, even if they partially reject the hypothesis with phrases like "not all" or "not always" or even outright deny it, they sometimes mention other external conditions or observational errors that influence the observation fact. 

According to Table \ref{tab:main_result}, the T5 models, which had a PRR in the range of 10\%, often generated responses that directly negated the hypothesis. Additionally, the T5 models produced a higher number of nonsensical responses such as "No," or indecisive responses like "It is impossible to tell," compared to decoder-only models (See Table \ref{tab:generation_statistics} in Appendix \ref{subsec:response_statistic}). The Llama2 series stood out among nine models in the generation task for crafting the most holistic arguments. They not only explained observation facts using other conditions and factors but also mentioned the possibility of measurement errors (see Appendix \ref{subsec:llama_response}). Furthermore, Llama2 models often initially negate the hypothesis but then mention that the situation occurred under different conditions or that the observation cannot be generalized (see Table \ref{tab:llama_2} in Appendix \ref{apdx:qualitative}). They sometimes generate answers arguing that we should not generalize the specific cases (See Table \ref{tab:llama_4} in Appendix \ref{apdx:qualitative}).   Conversely, Phi-2 and the GPT series tended to interpret the given holistic argument prompts more as logical problems than from a holistic view.  As the hypothesis is contradicted to the observation, these models follow the observation and refute the hypothesis. For example, faced with ``I observed that some cardboard materials are magnetic.'' against ``cardboard is always nonmagnetic,'' they concluded, ``The conclusion is that the initial hypothesis is incorrect. Not all cardboard is nonmagnetic,'' providing a logically perfect answer that explicitly addresses the contradiction.  Particularly, GPT-3.5 generates conservative responses stating that the observation does not support the hypothesis without determining the truth of the hypothesis (See Table \ref{tab:gpt_5} in Appendix \ref{apdx:qualitative}). More representative examples and qualitative analysis can be found in the Appendix \ref{apdx:qualitative}.

\begin{figure*}[t]

  \centering

  \includegraphics[width=\textwidth]{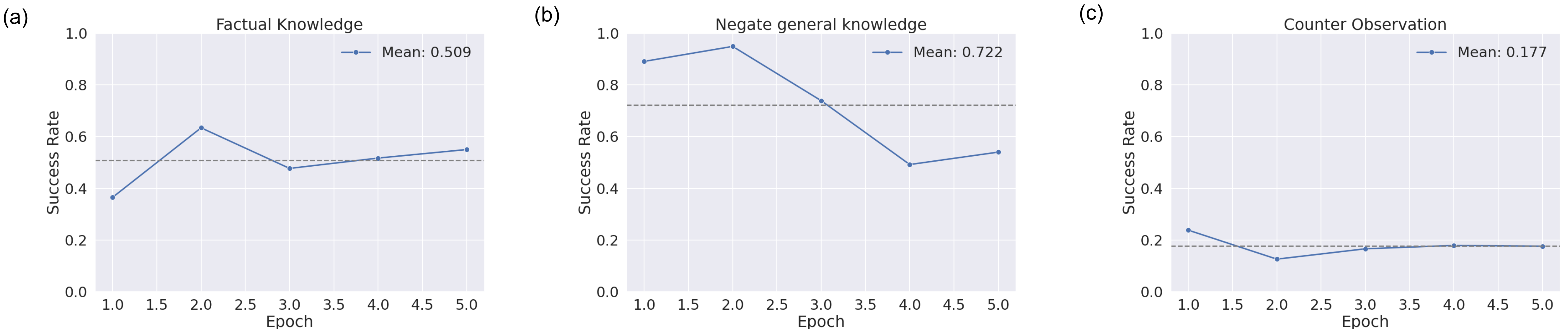}

 \captionsetup{}

  \caption{Changes in the success rate of knowledge edition over epochs during Knowledge Edit Supervised-Finetuning. (a) involves training on the negation of factual knowledge, while (b) and (c) involve fine-tuning the negation of general knowledge and the counter-observation of general knowledge, respectively. The success rate in (a) is the proportion at which the model negates the trained factual knowledge, and in (b) and (c), it is the rate at which general knowledge is answered as false.}

  \label{fig:sft}

\end{figure*}

\paragraph{Categorization of Peripheral Statements}

The dataset can be divided into two categories based on the peripheral statement $e^{*}_{2}$:

\begin{enumerate}
    \item $e^{*}_{2}$ that denies the observation itself by mentioning the atypicality of the observation.
    \item $e^{*}_{2}$ that explains the observation by using other factors or external conditions.
\end{enumerate}
We consider the former (1) to be a more immediate and more peripheral response to the counter observation of general knowledge. We randomly sampled 125 instances $(s,c,e^{*}_{1},e^{*}_{2})$ including $e^{*}_{2}$ that deny the observation fact and 125 instances including $e^{*}_{2}$ using external conditions. We then compared their performance in abduction and revision tasks. If LLMs impose a hierarchy on knowledge, considering the denial of an observation to be more peripheral, then the PRR for samples that deny the observation would be higher. However, the results in Figure \ref{fig:observ_etc} showed that the PRR was higher for samples introducing external conditions for both abduction and revision tasks. We interpret this as language models not being accustomed to denying the given context and thus having some difficulty in maintaining a holism network.

\begin{figure}[t]
    
    \centering
    
    \includegraphics[width=\columnwidth]{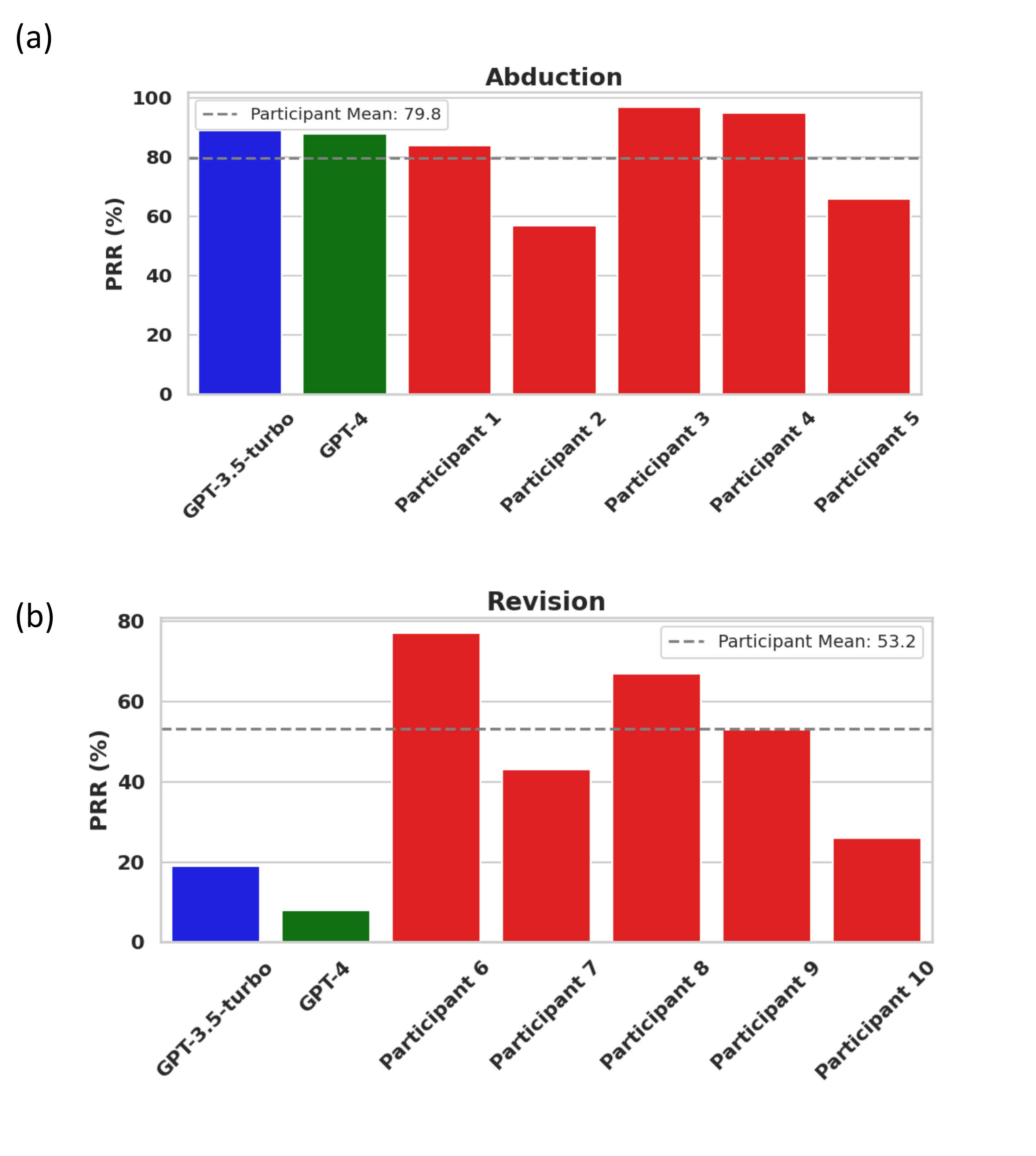}
    
    \captionsetup{}
    
    \caption{PRR score comparison between model and human responses. Different participants took part in each abduction and revision task. While GPT-3.5-turbo and GPT-4 exhibit similarities with human PRR in the abduction task, they significantly fall behind in the revision task.}
    
    \label{fig:human_eval}

\end{figure}

\paragraph{Can LLMs Protect Core Knowledge from Fine-Tuning?}
So far, we have explored whether language models can make holistic inferences about input contexts in a zero-shot manner. However, in this final experiment, we investigated whether language models are reluctant to modify core knowledge through supervised fine-tuning (SFT). Specifically, we tried to see if the language model still considers the target sentence $S$ to be true after being trained on $N$, the negation of the target sentence. We conducted three experiments with GPT-2, the first utilizing the factual knowledge in the FEVEROUS dataset \citep{aly2021feverous}, with the remaining two using datasets we created. In these three cases, N represents the negation of the FEVEROUS dataset, direct negation of S, and counter observation of S, respectively, and the success rate is the proportion of revision success, i.e., the rate at which the target sentence was answered as ``False.'' As can be seen (c) in Figure \ref{fig:sft}, after 5 epochs, when general knowledge was indirectly negated by counter-examples, the language model preserved its knowledge. However, merely indirectly negating does not allow us to ascertain if LLMs consider general knowledge as a core statement, as \citep{cohen2023evaluating} indicates vulnerability in knowledge editing to implicit consequences. Moreover, the final success rates in the two cases of direct negation were similar by (a), (b) in \ref{fig:sft}. In fact, the value was higher in the early stages when general knowledge was negated. Thus, through SFT, we can easily modify general knowledge itself, and models readily revise the general knowledge.

\paragraph{Human Evaluation}
We conducted a human evaluation with approval from the Institutional Review Board (IRB), involving a total of 10 participants. To mitigate potential cross-task influence, five participants were randomly assigned to solve the abduction task, and the other five tackled the revision task. We randomly sampled 100 questions each from a pool of 1709 abduction and revision problems, where participants had to choose the appropriate option for a given counter observation, same with the original tasks. As shown in Figure \ref{fig:human_eval}, for the abduction task, all participants demonstrated a PRR above 50\%, with two out of five correctly explaining the observed phenomena using peripheral statements in over 95 of the 100 sampled problems. For the revision task, among the participants, 3 out of 5 scored beyond a 50\% PRR, indicating that these individuals tend to revise peripheral knowledge rather than core knowledge when encountering abnormal situations. The remaining participants also had a stronger tendency to preserve core knowledge than the language models. You can see more detailed information in Appendix \ref{apdx:human_eval}.

\section{Conclusion}
The purpose of the paper was to explore whether LLMs exhibit characteristics consistent with epistemological holism, as they suggest that commonsense, general, and certain knowledge serves as the foundation of our belief network and is difficult to revise easily. Our findings reveal a nuanced picture: LLMs adeptly navigate peripheral beliefs in abduction tasks without negating core beliefs, showing proficiency in generating abductive explanations. However, across other evaluative frameworks, they exhibit a lesser tendency to recognize and protect the special status of core knowledge, suggesting a divergence from holism's principled knowledge interconnected hierarchy. The variability in results suggests that the extent to which language models conform to a holistic epistemological model varies by context, demonstrating an incomplete commitment to the principles of holistic epistemology. Consequently, it seems that language models do not consistently apply a holistic methodology to knowledge processing and reasoning across diverse cognitive challenges.

\section{Limitations and Further Research}
While we have explored the knowledge possessed by LLMs from the perspective of epistemological holism in this paper, it is important not to conclude from our results that language models possess an incorrect epistemology. Holism, though influential and plausible within the realms of linguistics and the philosophy of science, is but one among many theories of epistemology. Moreover, as we have discussed, how to form a priori interconnections and hierarchies among the knowledge within language models remains an open question for further inquiry. In this research, as our first aim is about the zero-shot inference of LLMs, we only conducted simple SFT experiments, but larger-scale experiments that influence core knowledge in different ways and examine its effects would also be necessary.

\section{Ethics Statement}
In the preparation of this paper, we utilized ChatGPT, for grammatical corrections and coding assistance. This technology served as an auxiliary resource to enhance the clarity and accuracy of our work, without directly influencing the research outcomes or decision-making processes involved. We acknowledge the support provided by OpenAI's ChatGPT in refining the presentation of our findings, ensuring that our use of this tool adheres to ethical guidelines and does not compromise the integrity of our research. 

\bibliography{anthology,custom}
\bibliographystyle{acl_natbib}

\newpage
\appendix

\onecolumn

\section*{Appendix}

\bigbreak
\section{Dataset Statistics}
The total number of samples in table $W$ is 9,727. By the extraction stage, we extracted 1,779 samples and filtered 7,948 out. After GPT`s overgeneration, we deleted 70 samples that contained repetitive sentences. 

\begin{table}[hbt!]

\centering
\small
\begin{tabular}{l|l|l|l|}
Source               & Counts & Source                      & Counts \\ \hline \hline
Action               & 179    & Dominant-recessiv           & 26     \\
Affect               & 50     & Environment                 & 18     \\
Affordances          & 27     & Hardness                    & 12     \\
Cause                & 181    & Inherited-learned           & 40     \\
Change               & 85     & Magnetism                   & 59     \\
Comparison           & 65     & Durability                  & 4      \\
Contains             & 99     & Opacity                     & 8      \\
Coupled relationship & 199    & Purity-mixture              & 9      \\
Durations            & 1      & Orbital                     & 7      \\
During               & 30     & Relative-distance           & 13     \\
Formedby             & 70     & Solubility                  & 6      \\
Frequency            & 7      & Shape-Volume                & 4      \\
Habitat              & 12     & State-of-matter-Temperature & 5      \\
Ifthen               & 171    & Things                      & 130    \\
Instances            & 3      & Warm-Cold blooded           & 7      \\
Lifespan             & 1      & Requires                    & 38     \\
Locations            & 37     & Sourceof                    & 1      \\
Process-stage-order  & 13     & Stage-in-Process            & 14     \\
Animal-reproduction  & 10     & Transfer                    & 16     \\
Chemical-charge      & 5      & UsedFor                     & 3      \\
Chemical-reaction    & 6      & Vehicle                     & 23     \\
Conductivity         & 15     & \textbf{Total}              & \textbf{1709}  
\end{tabular}%

\caption{Statistics for holism dataset.It consists of various fields of science.}
\label{tab:data_statistics}
\end{table}

\bigbreak
\clearpage

\section{Prompt Template}

\begin{table}[htb]
\begin{tcolorbox}[toprule=1mm,colback=gray!10!white,colframe=gray!50,title=Abduction Prompt Template, fonttitle=\bfseries]
\textbf{Observation}: \{observation\} \\
\textbf{Options}: \\
\textbf{(A)} \{explanation1\} \\
\textbf{(B)} \{explanation2\} \\
What is the best explanation for the observation? You should choose one among the options.
\end{tcolorbox}
\caption{The default prompt template for abduction task. \{observation\} refers to the counter-observation situation of the general scientific fact. \{explanation1\} and \{explanation2\} are randomly assigned either a negation of core knowledge or an explanation using a peripheral statement. }
\label{tab:abduction_prompt_template}
\end{table}

\begin{table}[htb]
\begin{tcolorbox}[toprule=1mm,colback=gray!10!white,colframe=gray!50,title=Revision Prompt Template, fonttitle=\bfseries]
\textbf{Observation}: \{observation\} \\
\textbf{Options}: \\
\textbf{(A)} \{explanation1\} \\
\textbf{(B)} \{explanation2\} \\
Considering the observation, which option is more likely to be revised? You should choose one among the options.
\end{tcolorbox}
\caption{The default prompt template for revision task. \{observation\} refers to the counter-observation situation of the general scientific fact. \{explanation1\} and \{explanation2\} are randomly assigned either a  core knowledge or explanation of the absence of other conditions or observation errors.}
\label{tab:abduction_prompt_template}
\end{table}

\bigbreak
\clearpage

\begin{table}[htb]
\begin{tcolorbox}[toprule=1mm,colback=gray!10!white,colframe=gray!50,title=Argument Generation Prompt Template, fonttitle=\bfseries]
\textbf{Hypothesis}: {hypothesis} \\
\textbf{Observation}: {observation} \\
What is the conclusion of given hypothesis and observation?
\end{tcolorbox}
\caption{The default prompt template for revision task. \{hypothesis\} refers to the the general scientific fact (core knowledge). {observation} is the counter-observation situation of the general scientific fact..}
\label{tab:abduction_prompt_template}
\end{table}

\bigbreak
\clearpage

\bigbreak
\clearpage
\section{Qualitative Analysis}\label{apdx:qualitative}
This section includes a detailed qualitative analysis of generation tasks. As the response style is different from the language models, we divided the response by each language model.
\subsection{Response Statistics}\label{subsec:response_statistic}

\begin{table}[h]

\centering
\small
\begin{tabular}{l|ll} \hline
Model           & \multicolumn{2}{c}{Argument Generation}                                              \\ \hline\hline
Metric          & \multicolumn{1}{l|}{Non-sensical or Ambiguous Response} & Withdraw Response \\ \hline
Flan-T5-Base    & \multicolumn{1}{l|}{59\%}                               & 3.7\%             \\ \hline
Flan-T5-Large   & \multicolumn{1}{l|}{55.1\%}                             & 5.9\%             \\ \hline
Flan-T5-XL      & \multicolumn{1}{l|}{58\%}                               & 3.1\%             \\ \hline
Flan-T5-XXL     & \multicolumn{1}{l|}{46\%}                               & 1.8\%             \\ \hline
Llama2-7b-chat  & \multicolumn{1}{l|}{3\%}                                & 0\%               \\ \hline
Llama2-13b-chat & \multicolumn{1}{l|}{2.2\%}                              & 0\%               \\ \hline
Phi-2           & \multicolumn{1}{l|}{3\%}                                & 0\%               \\ \hline
GPT-3.5-turbo   & \multicolumn{1}{l|}{11.3\%}                             & 0\%               \\ \hline
GPT-4           & \multicolumn{1}{l|}{1.7\%}                              & 0\%              \\ \hline
\end{tabular}%
\caption{The Non-sensical or Ambiguous Response refers to cases where the answer to a question is nonsensical, such as "No" or code generation. Withdraw Response refers to instances where the question is abandoned, as seen in responses like "It is impossible to tell." Withdraw Responses were primarily observed in T5 models.}
\label{tab:generation_statistics}
\end{table}

\bigbreak
\clearpage

\subsection{Responses from T5}\label{subsec:t5_response}

\begin{table}[htb]
\begin{tcolorbox}[toprule=1mm,colback=gray!10!white,colframe=gray!50,title=Negative PRR Response, fonttitle=\bfseries]
\textbf{Hypothesis}: an animal requires enough nutrients to maintain good health; survive. \\
\textbf{Observation}: I observed an animal that survived and maintained good health despite not receiving enough nutrients. \\
What is the conclusion of given hypothesis and observation? \\
\textbf{Answer}: An animal can survive and maintain good health without receiving enough nutrients.
\end{tcolorbox}
\caption{The case that the T5 model negates the core statement "I observed an animal that survived and maintained good health despite not receiving enough nutrients." by accepting the observation and making a hasty generalization.}
\label{tab:t5_1}
\end{table}

\begin{table}[htb]
\begin{tcolorbox}[toprule=1mm,colback=gray!10!white,colframe=gray!50,title=Negative PRR Response, fonttitle=\bfseries]
\textbf{Hypothesis}: the cocoons being created occurs during the the pupa stage in a life cycle.\\
\textbf{Observation}: I observed that cocoons are not always created during the pupa stage in a life cycle. \\
What is the conclusion of given hypothesis and observation? \\
\textbf{Answer}: It is not possible to tell
\end{tcolorbox}
\caption{The example of withdraw response of T5 model. The withdraw response only occurs in T5 models.}
\label{tab:t5_2}
\end{table}

\begin{table}[htb]
\begin{tcolorbox}[toprule=1mm,colback=gray!10!white,colframe=gray!50,title=Positive PRR Response, fonttitle=\bfseries]
\textbf{Hypothesis}: waves contain mechanical energy. \\
\textbf{Observation}: I observed a wave that did not contain mechanical energy. \\
What is the conclusion of given hypothesis and observation? \\
\textbf{Answer}: It is impossible to observe a wave that did not contain mechanical energy.
\end{tcolorbox}
\caption{The case that T5 model denies the observation by negating the observation.}
\label{tab:t5_3}
\end{table}

\bigbreak
\clearpage

\subsection{Responses from Llama}\label{subsec:llama_response}

\begin{table}[htb]
\begin{tcolorbox}[toprule=1mm,colback=gray!10!white,colframe=gray!50,title=Negative PRR Response, fonttitle=\bfseries]
\textbf{Hypothesis}: mice give birth to live young.\\
\textbf{Observation}: I observed a mouse laying eggs instead of giving birth to live young. \\
What is the conclusion of given hypothesis and observation? \\
\textbf{Answer}: Based on the given hypothesis and observation, the conclusion would be that the hypothesis is false.
The hypothesis states that mice give birth to live young, but the observation shows that the mouse laid eggs instead of giving birth to live young. This contradicts the hypothesis and suggests that mice do not give birth to live young.
Therefore, the conclusion is that the hypothesis is false and the observation provides evidence to support this conclusion.
\end{tcolorbox}
\caption{The case that Llama model denies general scientific fact and made a hasty generalization "mice do not give birth to live young."}
\label{tab:llama_1}
\end{table}

\begin{table}[htb]
\begin{tcolorbox}[toprule=1mm,colback=gray!10!white,colframe=gray!50,title=Positive PRR Response, fonttitle=\bfseries]
\textbf{Hypothesis}: an image in a mirror is formed by reflecting light.\\
\textbf{Observation}: I observed that an image in a mirror is not formed by reflecting light. \\
What is the conclusion of given hypothesis and observation? \\
\textbf{Answer}: The conclusion of the hypothesis and observation is that the image in a mirror is not formed by reflecting light.

The hypothesis states that an image in a mirror is formed by reflecting light. However, the observation contradicts this hypothesis by showing that an image in a mirror is not formed by reflecting light. \textbf{Therefore, the conclusion is that the image in a mirror is formed by some other means, such as by the brain's ability to process visual information and create a mental image of what is seen in the mirror, or by some other physical or psychological mechanism.It is important to note that the observation does not necessarily disprove the hypothesis entirely, but rather suggests that there may be other factors at play in the formation of images in mirrors. Further investigation and experimentation may be necessary to fully understand the mechanism by which images are formed in mirrors.}
\end{tcolorbox}
\caption{The example of Llama's Positive PRR case. Although the model negates the hypothesis at first, it generates the explanation by exploiting other conditions and factors that might affect the hypothesis and observations.}
\label{tab:llama_2}
\end{table}

\begin{table}[htb!]
\begin{tcolorbox}[toprule=1mm,colback=gray!10!white,colframe=gray!50,title=Positive PRR Response, fonttitle=\bfseries]
\textbf{Hypothesis}: electrons have the least mass in an atom.\\
\textbf{Observation}: I observed that protons have the least mass in an atom.\\
What is the conclusion of given hypothesis and observation? \\
\textbf{Answer}: \ldots \textbf{Therefore, the conclusion that logically follows from your hypothesis and observation is that you have made an error in your observation or reasoning. It is important to ensure that the conclusions we draw are based on accurate and consistent information to avoid drawing false conclusions.} \ldots
\end{tcolorbox}
\caption{The example of Llama's Positive PRR case. Instead of negating the hypothesis, the model mentions the possibility of errors in the observation.}
\label{tab:llama_3}
\end{table}

\begin{table}[H]
\begin{tcolorbox}[toprule=1mm,colback=gray!10!white,colframe=gray!50,title=Positive PRR Response, fonttitle=\bfseries]
\textbf{Hypothesis}: a desert environment usually is hot in temperature.\\
\textbf{Observation}: I experienced a desert environment that was cold in temperature.\\
What is the conclusion of given hypothesis and observation? \\
\textbf{Answer}: \ldots \textbf{In other words, the observation contradicts the hypothesis, so we cannot confirm the hypothesis as universally true. It is important to note that this conclusion is based on a single observation and may not be representative of all desert environments. Further research or additional observations may be necessary to confirm or refute the hypothesis.} \ldots
\end{tcolorbox}
\caption{The example of Llama's Positive PRR case. The response mentions that the counter-observation cannot be representative of all cases.}
\label{tab:llama_4}
\end{table}

\bigbreak
\clearpage

\subsection{Responses from Phi-2}\label{subsec:phi2_response}

\begin{table}[htb]
\begin{tcolorbox}[toprule=1mm,colback=gray!10!white,colframe=gray!50,title=Negative PRR Response, fonttitle=\bfseries]
\textbf{Hypothesis}: a tadpole lives in water.\\
\textbf{Observation}: I observed a tadpole living outside of water. \\
What is the conclusion of given hypothesis and observation? \\
\textbf{Answer}: The conclusion is that the tadpole is not living in water.
\end{tcolorbox}
\caption{The example where Phi-2 negates the core statement by accepting the observation.}
\label{tab:phi_1}
\end{table}

\begin{table}[htb]
\begin{tcolorbox}[toprule=1mm,colback=gray!10!white,colframe=gray!50,title=Positive PRR Response, fonttitle=\bfseries]
\textbf{Hypothesis}: reproductive behavior is an inherited characteristic.\\
\textbf{Observation}: I observed that reproductive behavior can vary within a species, even among individuals with the same genetic inheritance. \\
What is the conclusion of given hypothesis and observation? \\
\textbf{Answer}: The conclusion is that reproductive behavior is not solely determined by genetic inheritance, but can also be influenced by environmental factors and individual experiences.
\end{tcolorbox}
\caption{The case that T5 model denies the observation by negating the observation.}
\label{tab:phi_2}
\end{table}

\begin{table}[htb]
\begin{tcolorbox}[toprule=1mm,colback=gray!10!white,colframe=gray!50,title=Negative PRR Response, fonttitle=\bfseries]
\textbf{Hypothesis}: using tools is a learned characteristic.\\
\textbf{Observation}: I observed a group of animals using tools without any prior learning or training. \\
What is the conclusion of given hypothesis and observation? \\
\textbf{Answer}: The conclusion is that using tools is a learned characteristic.
\end{tcolorbox}
\caption{It is the case where the hypothesis is simply presented as the conclusion without explanation were also classified as positive.}
\label{tab:phi_3}
\end{table}

\bigbreak
\clearpage

\subsection{Responses from GPT}\label{subsec:gpt_response}

\begin{table}[htb]
\begin{tcolorbox}[toprule=1mm,colback=gray!10!white,colframe=gray!50,title=Negative PRR Response, fonttitle=\bfseries]
\textbf{Hypothesis}: circulation the circulatory system transports blood throughout the body.\\
\textbf{Observation}: I observed that the circulatory system does not transport blood throughout the body. \\
What is the conclusion of given hypothesis and observation? \\
\textbf{Answer}: The conclusion is that the hypothesis is incorrect.
\end{tcolorbox}
\caption{The case that GPT model directly negates the hypothesis by observing the counter example.}
\label{tab:gpt_1}
\end{table}

\begin{table}[htb]
\begin{tcolorbox}[toprule=1mm,colback=gray!10!white,colframe=gray!50,title=Positive PRR Response, fonttitle=\bfseries]
\textbf{Hypothesis}: the color of; coloration of fur is an inherited characteristic.\\
\textbf{Observation}: I observed that the coloration of fur in a litter of puppies was not consistent, despite both parents having the same fur color. \\
What is the conclusion of given hypothesis and observation? \\
\textbf{Answer}: The conclusion could be that the coloration of fur in puppies is not solely determined by the fur color of the parents, suggesting that other genetic factors or possibly recessive genes may be involved. This could potentially challenge the hypothesis that fur color is a straightforward inherited characteristic.
\end{tcolorbox}
\caption{The case that GPT model keeps the hypothesis by mentioning other possible factors. }
\label{tab:gpt_2}
\end{table}

\begin{table}[htb]
\begin{tcolorbox}[toprule=1mm,colback=gray!10!white,colframe=gray!50,title=Positive PRR Response, fonttitle=\bfseries]
\textbf{Hypothesis}: monkeys give birth to live young.\\
\textbf{Observation}: I observed a monkey laying eggs instead of giving birth to live young. \\
What is the conclusion of given hypothesis and observation? \\
\textbf{Answer}: Monkeys can lay eggs.
\end{tcolorbox}
\caption{The example discusses the same topic of inheritance as seen in Table \ref{tab:gpt_2} and explains the situation in the same manner.}
\label{tab:gpt_3}
\end{table}

\begin{table}[htb!]
\begin{tcolorbox}[toprule=1mm,colback=gray!10!white,colframe=gray!50,title=Positive PRR Response, fonttitle=\bfseries]
\textbf{Hypothesis}: the Law of Superposition is the lowest layers contain the oldest rocks in undisturbed layers of sedimentary rocks.\\
\textbf{Observation}: In a specific location, I observed that the lowest layers contain the youngest rocks in undisturbed layers of sedimentary rocks. \\
What is the conclusion of given hypothesis and observation? \\
\textbf{Answer}: The conclusion is that the given observation contradicts the hypothesis of the Law of Superposition. This could mean that the layers of rocks in that specific location have been disturbed or there is an error in the observation or understanding of the rock layers.
\end{tcolorbox}
\caption{GPT model denies the observation fact and protect the general hypothesis by mentioning the error in the observation}
\label{tab:gpt_4}
\end{table}

\begin{table}[H]
\begin{tcolorbox}[toprule=1mm,colback=gray!10!white,colframe=gray!50,title=Ambiguous Response, fonttitle=\bfseries]
\textbf{Hypothesis}: sunlight produces heat.\\
\textbf{Observation}: I observed that sunlight does not produce heat. \\
What is the conclusion of given hypothesis and observation? \\
\textbf{Answer}: The conclusion would be that the hypothesis is not supported by the observation.
\end{tcolorbox}
\caption{The response only states that the observation weakens the hypothesis without addressing whether the hypothesis is true or false, so it was classified as an ambiguous case. }
\label{tab:gpt_5}
\end{table}

\bigbreak
\clearpage

\section{Human evaluation}\label{apdx:human_eval}
This section includes a detailed analysis and template for human evaluation.
\subsection{Cohen`s Kappa between human and model responses }\label{subsec:cohen}

\begin{table}[hbt!]
\begin{tabular}{l|llllll}
\hline
\multicolumn{1}{l|}{}              & Participant 1 & Participant 2 & Participant 3 & Participant 4 & Participant 5 & Average \\ \hline\hline
\multicolumn{1}{l|}{GPT-3.5-turbo} & 0.106         & -0.126        & 0.1           & 0.06          & 0.07          & 0.042   \\ \hline
\multicolumn{1}{l|}{GPT-4}         & 0.338         & -0.141        & 0.09          & 0.177         & 0.154         & 0.123   \\ \hline
Phi-2                              & 0.402         & -0.072        & 0.19          & 0.135         & 0.165         & 0.164   \\ \hline

\end{tabular}%
\caption{Cohen`s Kappa for abduction task.}

\end{table}

\begin{table}[hbt!]
\begin{tabular}{l|llllll}
\hline
              & Participant 6 & Participant 7 & Participant 8 & Participant 9 & Participant 10 & Average \\ \hline \hline
GPT-3.5-turbo & -0.079        & -0.095        & 0.057         & 0.074         & -0.224         & -0.076  \\ \hline
GPT-4         & 0.023         & -0.02         & 0.051         & -0.009        & -0.005         & 0.008   \\ \hline
Phi-2         & -0.176        & -0.107        & 0.186         & 0.061         & -0.014         & -0.01   \\ \hline

\end{tabular}%
\caption{Cohen`s Kappa for revision task.}

\end{table}

\bigbreak
\clearpage

\subsection{Template for Human Evaluation}\label{subsec:human_temp}

\begin{figure}[hbt!]
    
    \centering
    
    \includegraphics[width=\textwidth]{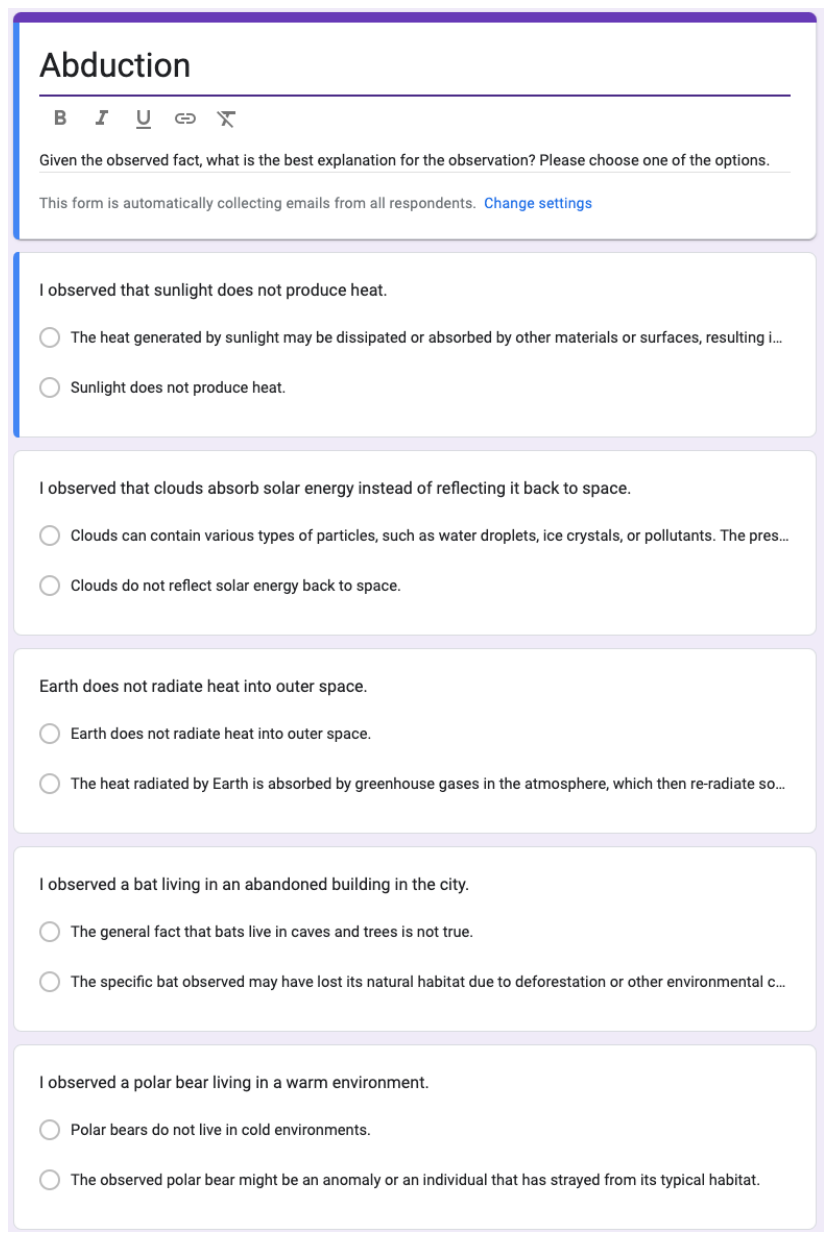}
    
    \captionsetup{}
    
    \caption{Google form template for abduction task.}
    
    \label{fig:human}

\end{figure}

\begin{figure}[hbt!]
    
    \centering
    
    \includegraphics[width=\textwidth]{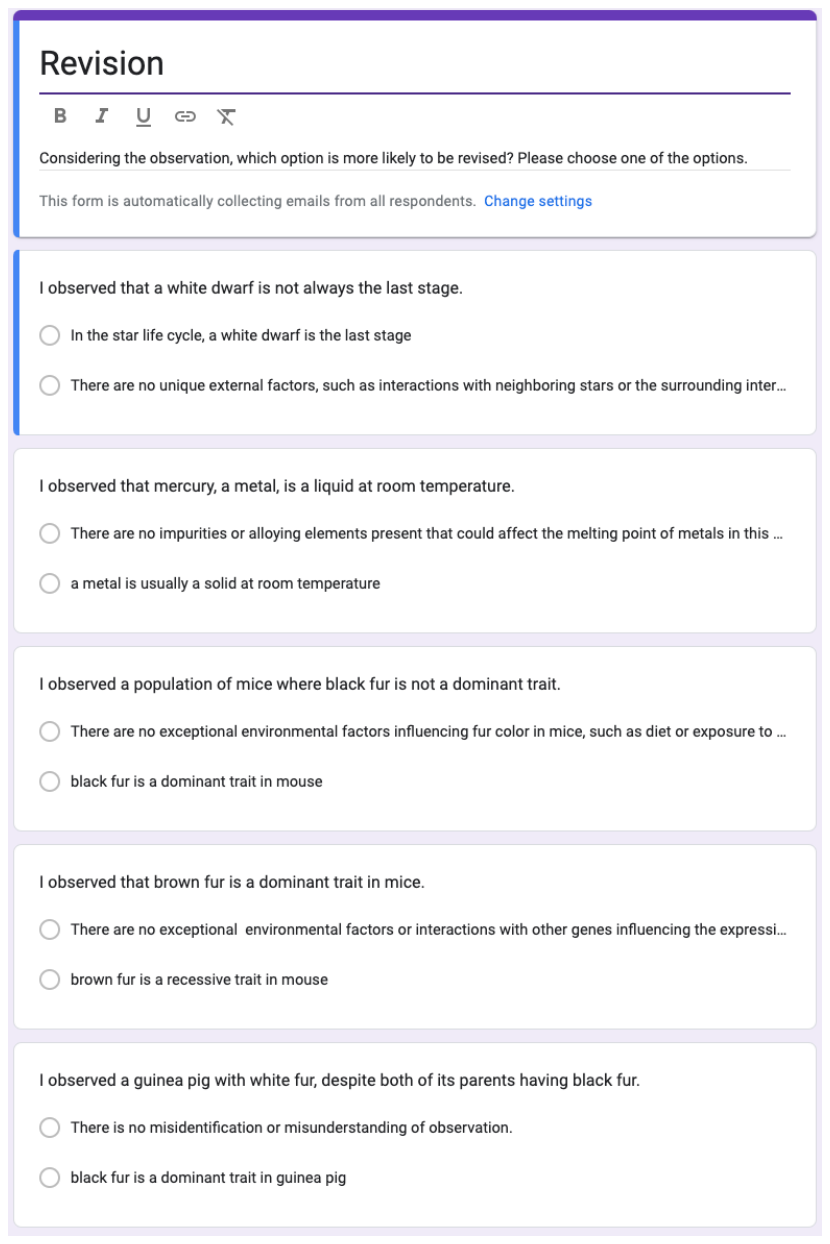}
    
    \captionsetup{}
    
    \caption{Google form template for revision task.}
    
    \label{fig:human}

\end{figure}

\end{document}